\documentclass[letterpaper, 10 pt, conference]{ieeeconf}
\IEEEoverridecommandlockouts

\makeatletter
\let\labelindent\@undefined
\makeatother
\providecommand{\tablename}{TABLE}

\usepackage{cite}
\usepackage{amsmath,amssymb,amsfonts}
\usepackage{graphicx}
\usepackage{textcomp}
\usepackage{xcolor}
\usepackage{bm}
\usepackage{booktabs,array,tabularx,multirow}
\usepackage{enumitem}
\usepackage{url}
\usepackage{dblfloatfix}
\usepackage{siunitx}
\usepackage[hidelinks,breaklinks=true]{hyperref}

\makeatletter

\def\@biblabel#1{[\raisebox{0pt}[0pt][0pt]{\hypertarget{Hcite.#1}{}}#1]}
\makeatother

\newif\ifanonymize
\anonymizefalse

\graphicspath{{figures/}}

\setlist[itemize]{leftmargin=1.2em,itemsep=1pt,topsep=2pt,parsep=0pt}

\newcommand{\mypara}[1]{\par\addvspace{1mm}\noindent\textbf{#1}}

\definecolor{GRLtodo}{HTML}{4e2f17}
\newif\ifshowtodos
\showtodostrue

\newcommand{\tablecaption}[1]{%
  \refstepcounter{table}%
  \begingroup\footnotesize
  \leftskip=0pt \rightskip=0pt \parfillskip=0pt plus 1fil
  \noindent\MakeUppercase{\tablename}~\thetable: #1\par
  \endgroup
  \vspace{0.5\baselineskip}%
}

\newcommand{\paperTitle}{The Cartesian Hand: In-Hand Manipulation with All-Linear Fingers}

\title{\paperTitle}

\ifanonymize
  \author{Anonymous Authors}
\else
  \author{Boxi Xia$^{\dagger}$, Bokuan Li$^{\dagger}$, Ryan Shin, Zijiang Yang, Jiaxun Liu, Boyuan Chen%
  \thanks{*All authors are from Duke University. $^{\dagger}$Equal contribution, co-first authors. This work is supported by DARPA FoundSci program under award HR00112490372, DARPA TIAMAT program under award HR00112490419, ARO under award W911NF2410405, ARL STRONG program under awards W911NF2320182, W911NF2220113, and W911NF242021.}}
\fi

\makeatletter
\let\IEEEaftertitletext\@IEEESAVECMDIEEEaftertitletext
\makeatother

\IEEEaftertitletext{%
  \vspace{-1.5\baselineskip}
  \begin{center}
    \includegraphics[width=\linewidth]{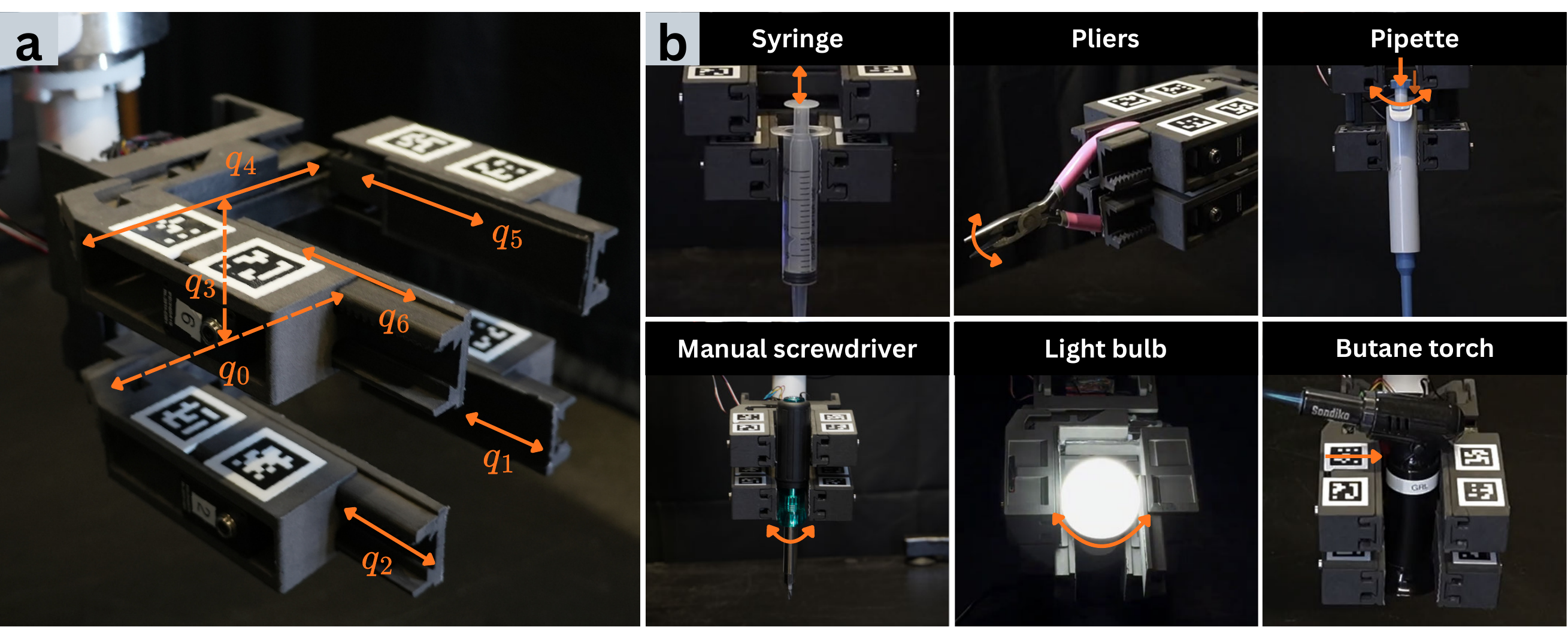}\\[4pt]
    \refstepcounter{figure}%
    \label{fig:teaser}%
    \parbox{\linewidth}{\footnotesize \textbf{Fig.~\thefigure.~The Cartesian hand executes in-hand manipulation with all-linear fingers.} \textbf{(a)}~Seven linear joints control gripper opening ($q_0,q_4$), gripper separation ($q_3$), and fingertip translation ($q_1,q_2,q_5,q_6$). \textbf{(b)}~Our hand handles a variety of challenging and articulated objects commonly found across laboratory, manufacturing, and household settings.}
  \end{center}
}

\begin{document}

\pagenumbering{arabic}
\setcounter{page}{1}

\maketitle

\begin{abstract}
Robotic manipulation has increasingly pursued human-like dexterous hands with many articulated degrees of freedom, offering rich manipulation capabilities at the cost of mechanical and control complexity. At the other extreme, parallel grippers are simple and robust, but provide little ability to manipulate an object after grasping it. Operating articulated objects such as threaded containers, manufacturing tools, and laboratory instruments often requires a second gripper, an external fixture, or coordinated arm motion. We introduce the \texttt{Cartesian Hand}, a 7-DoF end-effector that rethinks dexterous manipulation by combining independent grasping and relative manipulation within a single end-effector using only linear motion. Two independently actuated parallel grippers hold different parts of an object, while four translating fingertips generate relative motion between the grasped parts. Its configuration-independent fingertip kinematics allow manipulation to be composed from simple linear motion primitives. The Cartesian Hand is particularly suited to objects structured around common mechanisms such as threads, pivots, linear guides, plungers, and triggers. We demonstrate cap opening and closing, pipetting, pumping, two-handle manipulation, screwdriving, trigger actuation, and in-grasp reorientation across 35 objects spanning laboratory, manufacturing, and household settings. The same manipulation procedures transfer from a fixed-base robot arm to a humanoid, where we demonstrate bimanual laboratory manipulation using two Cartesian Hands. These results show that versatile in-hand manipulation capability can emerge from a mechanically simple architecture when independent grasping and relative motion are designed directly into the end-effector. 
We will open-source all software and hardware design.
Our website is \ifanonymize\url{https://anonymous.4open.science/w/cartesian-hand-web/}\else\url{https://generalroboticslab.com/cartesian_handv1}\fi.
\end{abstract}

\section{Introduction}
\label{sec:introduction}

Robots increasingly operate in environments designed around human hands, where interacting with an object often requires more than simply picking it up. Opening a bottle, using a screwdriver, operating a pipette, pulling a syringe, or squeezing a pair of pliers all require a robot to grasp an object while producing controlled motion between its parts. These capabilities are essential for robots to move beyond pick-and-place toward in-hand manipulation of tools, instruments, and everyday objects. Yet achieving such dexterous in-hand manipulation within a compact and reliable robot hand remains challenging~\cite{huang2025humanlike}.

Parallel grippers are widely used because they are simple, robust, and easy to control, but they provide limited ability to manipulate an object after grasping it. Tasks that require relative motion between object parts, such as turning a cap while holding its container or operating a two-handle tool, typically rely on a second gripper, environmental support, or arm re-grasping. This turns a local object-manipulation problem into one that depends on additional hardware, environmental contact, or coordinated arm motion.

A different approach is to increase the dexterity of the hand itself. Anthropomorphic and multi-fingered robotic hands introduce additional fingers and articulated joints that can generate rich in-hand motions and contact interactions~\cite{salisbury1982articulated}. Recent advances in teleoperation and learning have further expanded the behaviors that can be realized with these platforms~\cite{yang2025ace,he2024omnih2o,fu2024humanplus,ben2025homie}. This generality comes with considerably greater mechanical and control complexity, including many coupled degrees of freedom, configuration-dependent fingertip kinematics, and the need to coordinate multiple contacts~\cite{bicchi2000hands}. Other end-effectors seek additional in-hand capability while retaining fewer fingers, for example through active belts, rollers, wheels, compliant mechanisms, or specialized contact surfaces~\cite{yuan2020design,cai2023hand,xie2023hand,xiang2024adaptive,zheng2026mecanum}. These designs demonstrate that useful manipulation can be achieved without reproducing the full human hand, but often introduce additional mechanisms at the contact interface or focus on manipulating a single rigid object within the grasp. There remains an opportunity for an end-effector that preserves the mechanical and control simplicity of direct linear actuation while providing greater ability to manipulate an object after it has been grasped.

We introduce the \texttt{Cartesian Hand}, a 7-DoF end-effector that rethinks dexterous manipulation by combining independent grasping and relative manipulation within a single hand using only linear motion. The hand consists of two vertically stacked, independently actuated parallel grippers whose separation can also be controlled, with each of their four fingertips translating independently along a fixed axis. One gripper can therefore hold one part of an object while the other acts on another, and coordinated fingertip translations provide additional motions for turning, pushing, pulling, and reorienting the grasped parts. All seven actuated degrees of freedom are prismatic, giving a configuration-independent mapping between joint and fingertip motion. The desired controls can be constructed from simple linear motion primitives without requiring extensive optimization or human motion tracking.

A key advantage of our design is that relatively simple hand motions are sufficient to operate a broad class of articulated objects. Many of these objects already constrain the relative motion between their parts. For example, a cap follows a thread, pliers rotate about a pivot, a syringe plunger translates along its barrel, and a trigger moves along a constrained path. The Cartesian Hand provides independent grasps and relative actuation to operate these mechanisms without requiring the robot to reproduce the complex finger motions a human hand would use. Repeated release, reset, and re-grasp further allow the hand to execute motions, such as multi-turn rotation, that exceed the travel of any individual actuator. This makes objects built around common mechanisms such as threads, pivots, linear guides, plungers, and triggers particularly well suited to the Cartesian Hand.

We evaluate the Cartesian Hand on 35 objects spanning scientific laboratory, manufacturing, and household settings. A single hand can perform cap opening and closing, pipetting, pumping, manipulation of two-handle tools, screwdriving, trigger actuation, and in-grasp reorientation using the same hardware and motion primitives. Manipulation is executed using joint feedback for contact detection without visual signals. We further demonstrate that the same manipulation procedures transfer from a fixed-base Franka Emika Panda to a humanoid robot~\cite{xia2026workspace}. Finally, we equip the humanoid with two Cartesian Hands and demonstrate bimanual laboratory manipulation in which one hand independently opens a centrifuge tube while the other operates a pipette.

The main contributions of this work are:
\begin{itemize}
    \item We introduce the Cartesian Hand, a 7-DoF end-effector with two independently actuated grasps and relative object manipulation within a single all-linear mechanism.
    \item We characterize its configuration-independent fingertip kinematics and develop a set of reusable motion primitives through which coordinated linear actuation produces grasping, translation, rotation, and relative manipulation.
    \item We demonstrate a broad range of in-hand manipulation capabilities across 35 laboratory, manufacturing, and household objects, and show that the same manipulation procedures transfer across robot embodiments and compose for bimanual manipulation on a humanoid.
\end{itemize}

\section{Related Work}
\label{sec:related}

\mypara{Dexterous robotic hands.} Multi-fingered robotic hands achieve in-hand manipulation through articulated fingers and coordinated contacts~\cite{salisbury1982articulated,bicchi2000robotic}. Anthropomorphic platforms have enabled increasingly capable systems for teleoperation, imitation learning, and general-purpose manipulation~\cite{yang2025ace,he2024omnih2o,fu2024humanplus,ben2025homie}. Recent vision-language-action models further extend these capabilities through large-scale learned policies~\cite{kim2024openvla,black2024pi0}. These approaches combine highly articulated hardware with increasingly sophisticated control and learning. The Cartesian Hand explores a different route to in-hand dexterity, where carefully designed mechanics reduce the manipulation problem to simple, directly controlled motions without relying on highly articulated hand kinematics.

\mypara{In-hand manipulation with augmented grippers.} A complementary line of work extends simpler grippers by introducing mobility at the contact interface. Active-surface designs use belts, rollers, wheels, vibration, or compliant surfaces to translate and rotate objects within a grasp~\cite{yuan2020design,cai2023hand,xie2023hand,isobe2023vision,nahum2022robotic,xiang2024adaptive,zheng2026mecanum}. In particular, BACH uses actuated belts for manipulation within a power grasp~\cite{cai2023hand}, while BOP integrates independently driven belts into a parallel-jaw gripper to produce motion primitives for object reorientation and tool use~\cite{xie2023hand}. The Cartesian Hand introduces mobility differently. Its contact surfaces remain passive while the fingertips themselves translate, and two independently actuated grippers can hold and manipulate different parts of an object. This enables relative manipulation between two grasps within a single hand, a capability particularly useful for articulated objects.

\mypara{Manipulation of articulated objects.} Articulated-object manipulation has been studied through perception, articulation modeling, planning, and learned interaction policies~\cite{capitanelli2018manipulation,xiang2020sapien,mo2021where2act,zhang2023flowbot}. Related work on extrinsic dexterity has shown that simple grippers can achieve richer manipulation by exploiting gravity, arm motion, or environmental contacts~\cite{dafle2014extrinsic,zhou2023learning}. Our work addresses the complementary hardware problem of executing relative manipulation directly within a single end-effector. The Cartesian Hand independently grasps two parts of an articulated object and actuates their relative motion, reducing reliance on a second gripper, external fixtures, or environmental contacts to provide the opposing grasp. Since the object's mechanism already constrains the relative motion between its parts, the same simple linear motions can operate objects built around common mechanisms such as threads, pivots, linear guides, plungers, and triggers.

\begin{figure}[t]
\centering
\includegraphics[width=\linewidth]{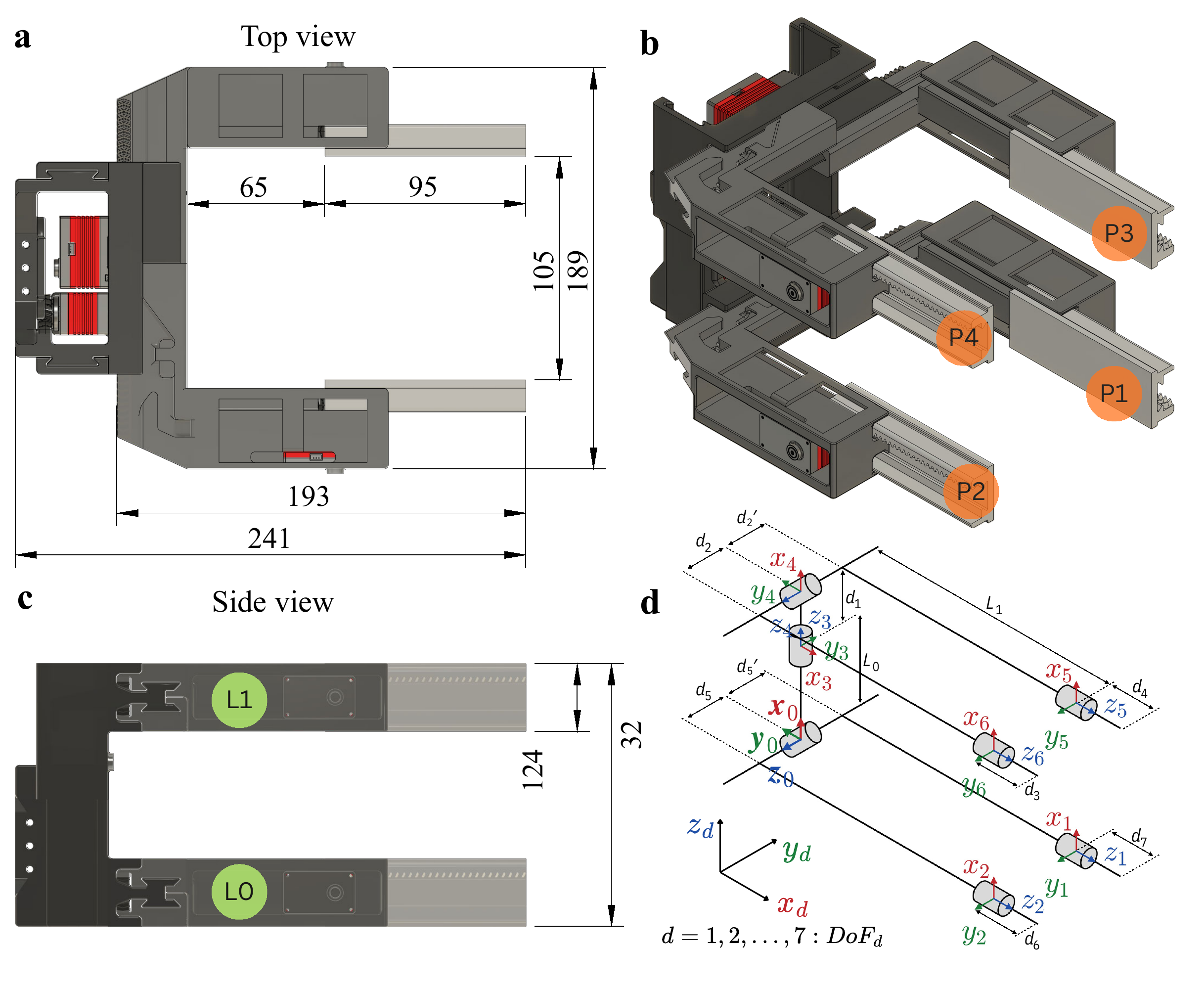}
\vspace{-20pt}
\caption{\textbf{Cartesian hand} with two stacked, identical parallel-jaw grippers, each with 2 sliding fingers. Green labels indicate the two actuated parallel grippers: L0~lower level (base) gripper, L1~upper level (auxiliary) gripper which can slide vertically relative to the base level gripper. Orange numbers indicate actuated fingertips: ($P_1$)~lower left fingertip, ($P_2$)~lower right fingertip, ($P_3$)~upper left fingertip, ($P_4$)~upper right fingertip. \textbf{(a)} Top view, \textbf{(b)} Isometric view, \textbf{(c)} side view, \textbf{(d)} joint kinematics. Dimensions in (a) and (c) are in mm.}
\vspace{-12pt}
\label{fig:ee}
\end{figure}

\section{Mechanical Design and Kinematics}
\label{sec:design}

\mypara{Design and Actuation.} The \texttt{Cartesian Hand} is a 7-DoF end-effector composed entirely of prismatic joints (Fig.~\ref{fig:ee}). Its architecture consists of two independently actuated parallel grippers arranged vertically, which we refer to as the base and auxiliary grippers. Each gripper contributes one DoF for opening and closing, and each of its two fingertips independently translates along the longitudinal direction of the finger. A seventh DoF translates the auxiliary gripper relative to the base gripper. These motions collectively provide independent grasping at two locations and relative translation between the two grasps, while the fingertip slides provide additional motion within each grasp.

We denote the seven joint coordinates by $\mathbf{q}=[q_0,q_1,q_2,q_3,q_4,q_5,q_6]^\top$ where $q_0$ and $q_4$ control the opening of the base and auxiliary grippers, respectively. $q_1,q_2$ and $q_5,q_6$ control their four fingertip slides. $q_3$ controls the separation between the two grippers. As illustrated in Fig.~\ref{fig:ee}, the fingertip slides move along the $x$-axis, the grippers open and close along the $y$-axis, and the auxiliary gripper translates along the $z$-axis. All actuated motion is therefore generated along fixed Cartesian directions.

This arrangement is designed around in-hand manipulation. The two grippers can grasp different parts of the same object and apply motion between them without requiring the robot arm to provide the opposing grasp. Within either gripper, coordinated fingertip translation can move or rotate a grasped part through contact. The separation DoF further produces relative motion between the two grasp locations. These capabilities can be combined while the arm remains stationary, allowing manipulation to occur locally within the end-effector.

All seven joints use rack-and-pinion transmissions driven by Feetech-3915 servos, with dovetail rails constraining the linear motion. The fingertip slides provide 65~mm of travel, while the gripper and inter-gripper separation axes each provide 52~mm of travel. Each joint reaches approximately 60~mm/s. The assembled hand weighs 850~g and measures approximately $166 \times 100 \times 76$~mm when fully closed. Each servo has a rated peak stall torque of 1.39~N$\cdot$m, corresponding to a theoretical peak rack force of approximately 170~N before transmission losses. We use friction tape on the fingertip to increase friction. In static testing, a single gripper can continuously hold a 2~kg bottle.

The hand is designed for low-cost fabrication and can be produced using fused-filament or selective-laser-sintering (SLS) additive manufacturing. The structural components cost approximately \$30 in PLA or \$50 in SLS Nylon 12, and the complete hand costs approximately \$500 including seven servos, electronics, wiring, and fasteners. The PLA version can be assembled in approximately two hours after printing. Both fabrication variants have been tested successfully for the manipulation tasks presented in this work.

\mypara{Fingertip Kinematics.} The all-prismatic architecture gives the Cartesian Hand a particularly simple relationship between actuator motion and fingertip motion. Let $\mathbf{p}_i \in \mathbb{R}^3$ denote a reference point fixed to fingertip $i$, with $\mathbf{p}_1$ and $\mathbf{p}_2$ belonging to the base gripper and $\mathbf{p}_3$ and $\mathbf{p}_4$ to the auxiliary gripper. Choosing the origin and joint-zero positions for notational convenience, their positions in the hand frame are
\begin{equation}
\begin{aligned}
\mathbf{p}_1 &=
\begin{bmatrix}
q_1 \\ -q_0 \\ 0
\end{bmatrix},
&
\mathbf{p}_2 &=
\begin{bmatrix}
q_2 \\ q_0 \\ 0
\end{bmatrix}, \\
\mathbf{p}_3 &=
\begin{bmatrix}
q_5 \\ -q_4 \\ q_3
\end{bmatrix},
&
\mathbf{p}_4 &=
\begin{bmatrix}
q_6 \\ q_4 \\ q_3
\end{bmatrix}.
\end{aligned}
\label{eq:fingertip_positions}
\end{equation}

Stacking the four reference points into $\mathbf{p} = [\mathbf{p}_1^\top,\mathbf{p}_2^\top, \mathbf{p}_3^\top,\mathbf{p}_4^\top]^\top$ gives $\mathbf{p} = \mathbf{J}\mathbf{q}, \qquad \mathbf{p}\in\mathbb{R}^{12}, \quad \mathbf{J}\in\mathbb{R}^{12\times7},$ where
\begin{equation}
\mathbf{J} =
\begin{bmatrix}
0& 1& 0& 0& 0& 0& 0\\
-1&0& 0& 0& 0& 0& 0\\
0& 0& 0& 0& 0& 0& 0\\
0& 0& 1& 0& 0& 0& 0\\
1& 0& 0& 0& 0& 0& 0\\
0& 0& 0& 0& 0& 0& 0\\
0& 0& 0& 0& 0& 1& 0\\
0& 0& 0& 0&-1& 0& 0\\
0& 0& 0& 1& 0& 0& 0\\
0& 0& 0& 0& 0& 0& 1\\
0& 0& 0& 0& 1& 0& 0\\
0& 0& 0& 1& 0& 0& 0
\end{bmatrix}.
\label{eq:jacobian}
\end{equation}
Unlike an articulated finger whose differential kinematics vary as its joints rotate, $\mathbf{J}$ is constant throughout the feasible workspace of the Cartesian Hand. Its seven columns are nonzero and have disjoint support, giving $\operatorname{rank}(\mathbf{J}) = 7$. The joint-to-fingertip velocity mapping retains full column rank throughout the feasible workspace and does not encounter configuration-dependent kinematic singularities.

\begin{figure}[t]
\centering
\includegraphics[width=\linewidth]{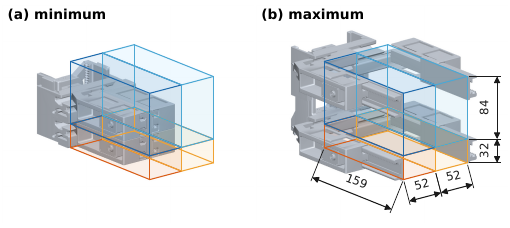}
\vspace{-24pt}
\caption{\textbf{Linear joints make each finger's reachable workspace an axis-aligned box.} The hand at its two end configurations, \textbf{(a)}~minimum configuration, and \textbf{(b)}~maximum configuration. The workspace of the base gripper is colored orange, and the workspace of the auxiliary gripper is colored blue. The two regions of a gripper meet on the closing midplane, where the finger faces touch when shut.}
\vspace{-12pt}
\label{fig:workspace}
\end{figure}

More importantly for manipulation, the same joint displacement always produces the same displacement of its associated fingertip reference points regardless of the current hand configuration. Manipulation behaviors can hence be specified directly as coordinated linear joint motions without solving configuration-dependent finger kinematics. This property provides the basis for the manipulation primitives introduced next. The same linear structure also produces an axis-aligned fingertip workspace (Fig.~\ref{fig:workspace}). Including fingertip dimensions and joint travel, the reachable region of each base-gripper fingertip spans approximately $159 \times 52 \times 32$~mm, while the auxiliary-gripper fingertips span approximately $159 \times 52 \times 84$~mm due to the additional inter-gripper translation.

\mypara{Manipulation Primitives.} A key benefit of the all-linear architecture is that its manipulation behaviors can be constructed systematically from a small number of coordinated actuator motions. We begin by treating each of the seven joints as either holding its position or moving toward a commanded position, giving $2^7=128$ possible binary actuation states. Most of these states are redundant because of the symmetry and coordinated operation built into the mechanism.

During normal manipulation, the two fingertip slides on each gripper are coordinated as a pair: $(q_1,q_2)$ for the base gripper and $(q_5,q_6)$ for the auxiliary gripper. Together with the two gripper joints $q_0$ and $q_4$ and the inter-gripper translation $q_3$, this gives five effective binary actuation groups, hence $2^5 = 32$ paired-motion states. We additionally retain four special cases in which a single fingertip moves independently, resulting in 36 candidate states. These four cases are kinematically equivalent and correspond to the same point-pushing/pulling behavior.

A second reduction follows from the symmetry between the two gripper levels. Exchanging the base and auxiliary grippers maps $(q_0,q_1,q_2,q_3,q_4,q_5,q_6) \longleftrightarrow (q_4,q_5,q_6,q_3,q_0,q_1,q_2),$ while preserving the corresponding manipulation behavior up to the choice of gripper role. Among the 32 paired-motion states, eight are unchanged by this exchange and the remaining 24 form 12 symmetric pairs, giving $8 + \frac{24}{2} = 20$ distinct classes. Treating the four equivalent single-fingertip cases as one additional class leaves 21 symmetry-reduced actuation patterns.

\begin{figure}[t]
\centering
\includegraphics[width=\linewidth]{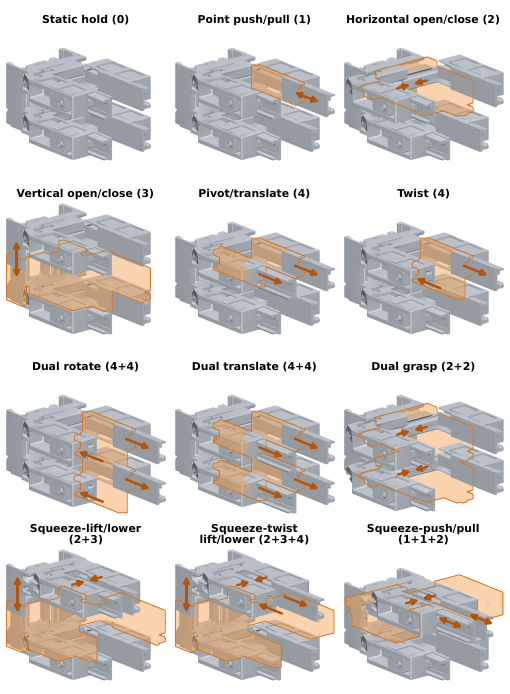}
\caption{\textbf{In-hand primitives for the Cartesian hand.} The moving components are marked with orange outlines, with arrows indicating the moving direction. }
\label{fig:primitives}
\end{figure}

\setlength{\tabcolsep}{3pt}
\begin{table}[t]
\centering
\tablecaption{Manipulation primitives of the \texttt{Cartesian Hand}. Motions that differ only by exchanging the base and auxiliary grippers are grouped into the same primitive family.}
\label{tab:primitives}
\begin{tabular}{c l l}
\hline
\textbf{Primitive} & \textbf{Representative DoFs} & \textbf{Function} \\
\hline
0 & -- & Static hold \\
1 & $q_1,q_2,q_5,$ or $q_6$ & Point push/pull \\
2 & $q_0$ or $q_4$ & Horizontal open/close \\
3 & $q_3$ & Vertical open/close \\
4 & $(q_1,q_2)$ or $(q_5,q_6)$ & Translate / pivot / twist \\
4+4 & $(q_1,q_2,q_5,q_6)$ & Dual rotate \\
4+4 & $(q_1,q_2,q_5,q_6)$ & Dual translate \\
2+2 & $q_0$ and $q_4$ & Dual grasp \\
2+3 & $(q_0,q_3)$ or $(q_3,q_4)$ & Squeeze-lift/lower \\
2+3+4 & $(q_0,q_1,q_2,q_3)$ or $(q_3,q_4,q_5,q_6)$ & Squeeze-twist-lift/lower \\
1+1+2 & $(q_0,q_1,q_2)$ or $(q_4,q_5,q_6)$ & Squeeze-push/pull \\

\hline
\end{tabular}
\vspace{-10pt}
\end{table}

These patterns can be simplified further because motions involving several simultaneously moving groups can be composed from simpler ones, and several symmetry-reduced states produce the same functional interaction with an object. Therefore, we select five reusable joint-action primitives and define six compositions of these primitives to support the manipulation tasks considered in this work (Table~\ref{tab:primitives}; Fig.~\ref{fig:primitives}). The reduction provides a compact and structured vocabulary from which the manipulation procedures can be constructed systematically.

The primitives expose how simple linear actuator motions produce more complex object motions. Closing the two grippers establishes independent grasps on different object parts, and translation of $q_3$ changes their relative position to produce pushing, pulling, or separation. Within either gripper, coordinated fingertip translation can translate, pivot, or twist a grasped part through contact, while combining gripper and separation motions produces additional behaviors such as squeeze-and-lift. These primitives are particularly effective for articulated objects, whose mechanisms already constrain the relative motion between their parts. For example, a threaded connection constrains the motion of a cap relative to its container, while a pivot constrains the motion of pliers or scissors. Objects with the same type of articulation can share the same primitive with different motion parameters. More complex procedures can be constructed by sequencing these primitives. A multi-turn cap combines dual grasping and fingertip twisting with repeated release, reset, and re-grasp. A syringe combines dual grasping with relative translation between the grippers. Thus, while every commanded actuator motion remains linear, their composition supports a broad range of translational and rotational in-hand manipulation.

\begin{figure*}[!t]
\centering
\includegraphics[width=\textwidth]{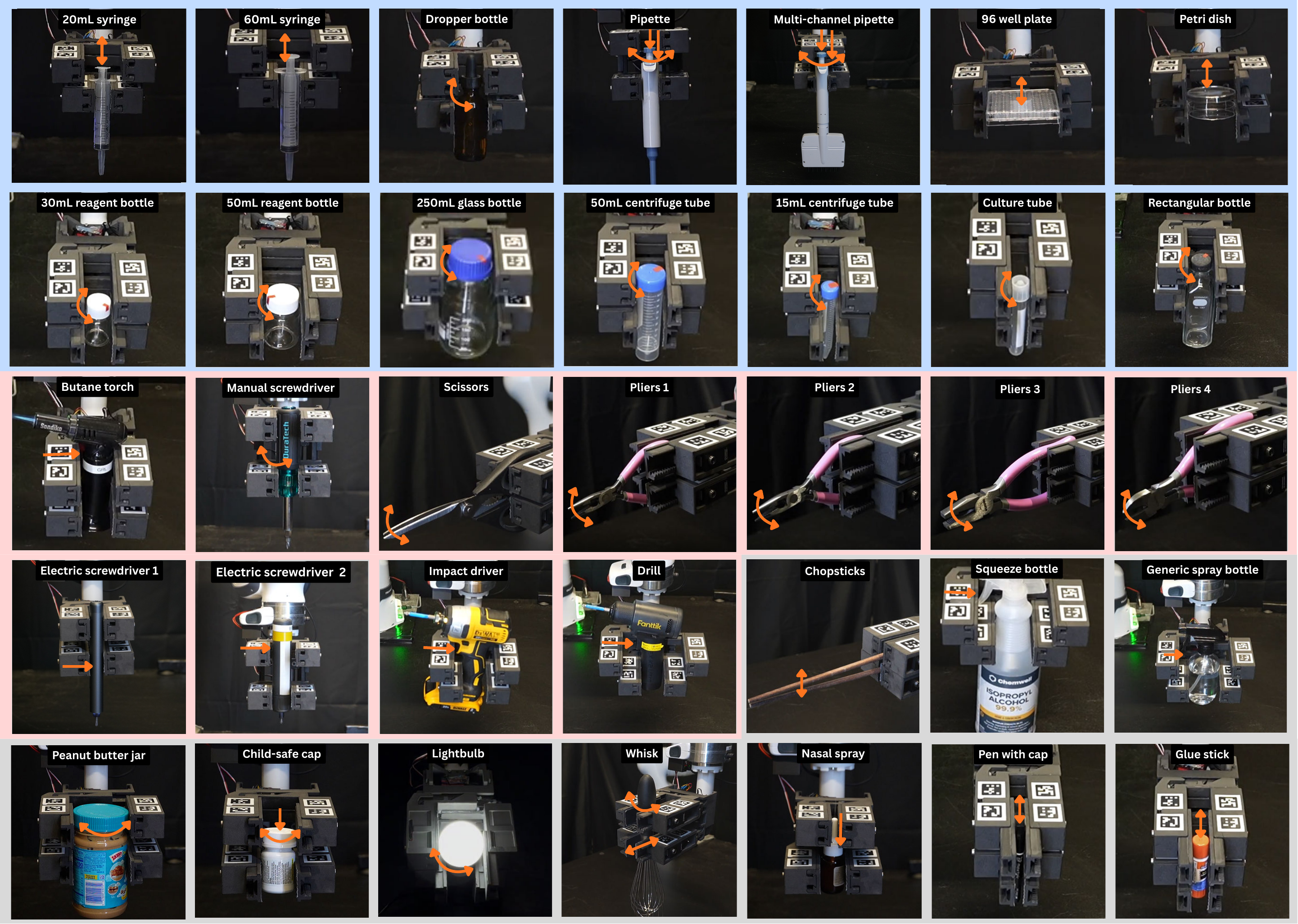}
\vspace{-20pt}
\caption{\textbf{The object set spans 35 objects across three domains.} The tested objects are grouped into laboratory, makerspace, and household categories. The set includes multiple sizes or variants of syringes, pipettes, pliers, and spray bottles. Orange arrows indicate object motion.}
\label{fig:objects}
\end{figure*}

\section{Control and Manipulation Procedures}
\label{sec:control}

\mypara{Contact-Aware Joint Control.} The Cartesian Hand executes manipulation using position-controlled joint motions and proprioceptive feedback from the servos. Each primitive specifies target positions, motion directions, speeds, and effort limits for its active joints. During grasping or contact-seeking motions, the controller detects contact when joint speed stays below a threshold for a set duration despite the joint being commanded to move with a force limit. The commanded motion is then terminated or held, and the servo joint position at contact is saved, allowing the same primitive to adapt to objects of different dimensions without requiring their geometry to be specified precisely.

This contact-based control also provides simple geometric information during manipulation. For example, when a gripper closes around a cap, its final jaw position estimates the cap diameter. The controller uses this estimate to determine the fingertip displacement and number of re-grasping cycles required for a requested rotation. Similar contact-aware motions allow the grippers to accommodate variations in object width. No visual feedback is used in this work. The Cartesian Hand includes embedded AprilTags, and we leave visual feedback for future work.

\mypara{Task Execution.} Each task is represented as a sequence of the manipulation primitives introduced above with a small set of object-specific parameters. These parameters specify quantities such as grasp offsets, separation between the two grasp locations, fingertip stroke distance, motion speed, and joint effort limits. Parameters that can be inferred from contact, such as object width or cap diameter, are measured automatically during execution. Offsets that cannot be determined from joint contact, such as the relative height of two object parts, are specified for each object.

Objects are initially presented at known poses in a calibrated workstation. The robot arm approaches the object using a predefined grasp pose, after which the hand establishes the required grasp and lifts the object from its holder. All subsequent manipulation is performed in hand, and the holder does not provide support or reaction force during the reported procedures. The robot arm remains stationary during the in-hand manipulation.

Objects sharing the same articulation use the same procedure with different parameter values. Threaded objects use repeated grasp, twist, release, and reset motions. Two-handle mechanisms and plungers use relative translation between the two grippers. Triggers use gripper closure followed by the object's return mechanism. Procedures can also be reversed where appropriate, allowing the same primitive sequence to both remove and reattach a threaded closure. To support a new object, we select the corresponding procedure and adapt only its scalar parameters, without changing the hand mechanism or the underlying primitive library. After establishing a procedure using one representative object from each category, adapting the procedure to a new object required at most five setup trials to determine the appropriate parameters in our experiments.

\begin{figure*}[!t]
\centering
\vspace{6pt}
\includegraphics[width=\textwidth]{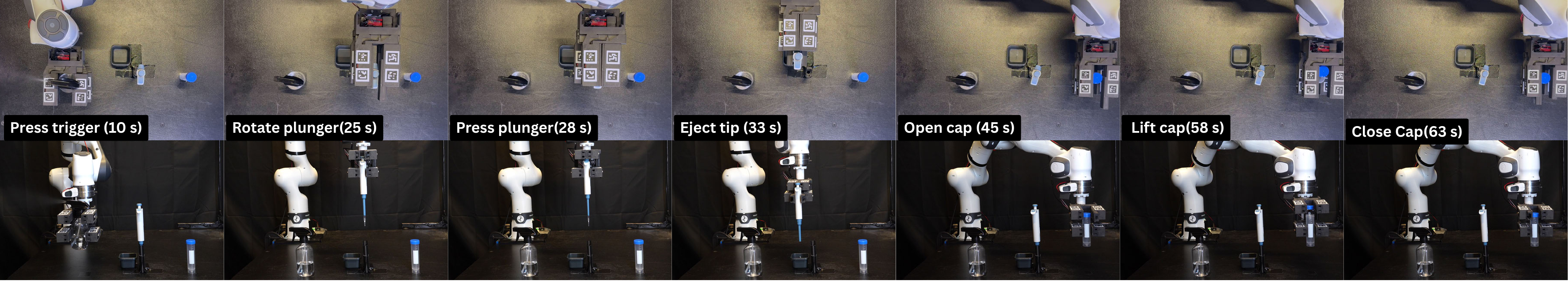}
\vspace{-20pt}
\caption{\textbf{A continuous sequence demonstrates manipulation along three actuation axes.} Two synchronized views of the sequence on the Franka embodiment, with the workstation top view above and the front view below: the hand pulls a spray bottle's trigger by closing its jaw, presses a pipette's plunger with the separation actuator, and twists a threaded cap off a bottle. The fingertip geometry remains unchanged throughout, and only the procedure settings change between objects. }
\label{fig:sequence}
\end{figure*}

\begin{figure*}[!t]
\centering
\includegraphics[width=\textwidth]{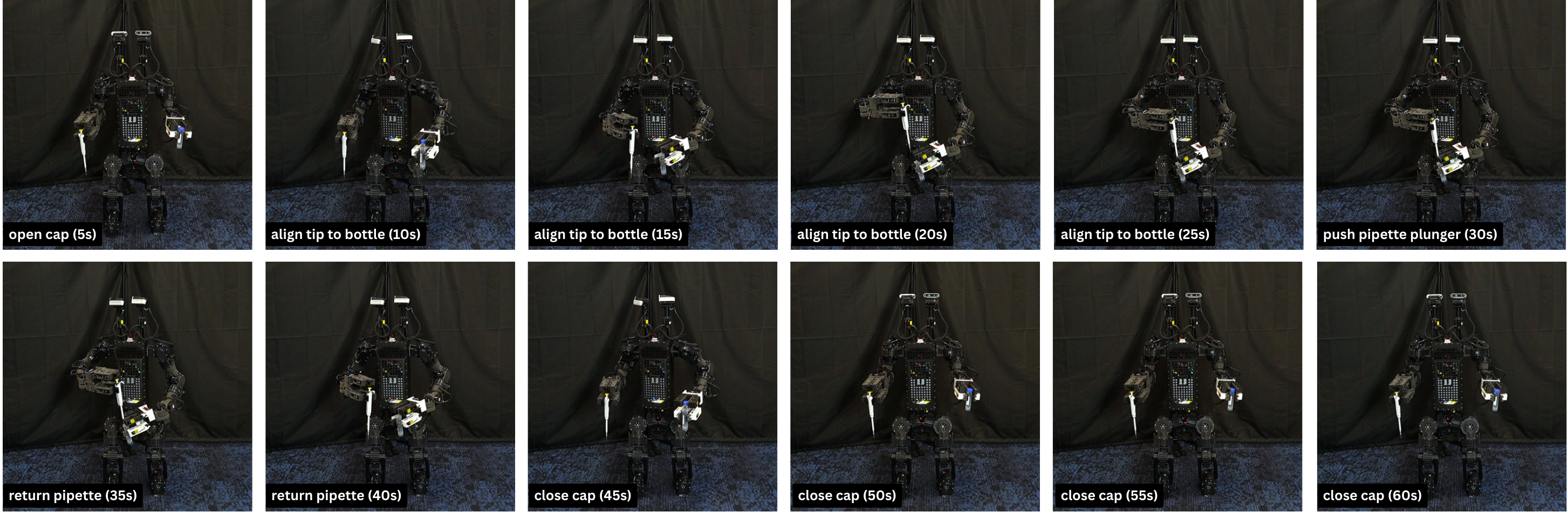}
\vspace{-20pt}
\caption{\textbf{Two Cartesian hands transfer liquid without coordinating on a shared object.} The humanoid holds a 50\,mL centrifuge tube in one hand and a pipette in the other. The tube-holding hand opens the tube on its own, its base gripper clamping the body while its auxiliary gripper twists off the cap; the pipetting hand then dispenses into the open tube. Each hand operates its own object, while the arms position the tube and pipette for liquid transfer.}
\label{fig:humanoid-sequence}
\end{figure*}

\section{Experiments}
\label{sec:experiments}

We evaluated the Cartesian Hand to answer four questions: (1) Can a small set of linear manipulation primitives operate a broad range of articulated objects? (2) How reliably can these procedures be executed across objects with different geometries? (3) Can the same manipulation procedures transfer across robot embodiments? (4) Can two hands compose their capabilities in a bimanual manipulation task?

\subsection{Experimental Setup}
\label{sec:experimental_setup}

We evaluated the Cartesian hand on 35 objects spanning scientific laboratory, manufacturing, and household settings (Fig.~\ref{fig:objects} and Table~\ref{tab:objects}). The set included threaded containers, syringes, single- and multi-channel pipettes, pumps, scissors, pliers, screwdrivers, triggers, and other commonly encountered objects. We organized these objects according to the type of manipulation required, including threaded rotation, pumping, pipetting, two-handle actuation, screwdriving, and trigger actuation.


Quantitative experiments were performed with the \texttt{Cartesian Hand} mounted on a 7-DoF Franka Emika Panda. Objects were placed at prescribed poses in a calibrated workstation and approached using predefined grasp poses. After the initial grasp, the object was lifted from its holder and all subsequent manipulation was performed in hand. The holder did not provide support or reaction force during manipulation. The arm remained stationary throughout the in-hand procedure.

\begin{table}[t]
\centering
\vspace{10pt}
\tablecaption{Object set spanning 35 objects across three domains and seven task categories.}
\label{tab:objects}
\small
\setlength{\tabcolsep}{3pt}
\renewcommand{\arraystretch}{1.05}
\begin{tabularx}{\columnwidth}{@{}llX@{}}
\toprule
Domain & Category & Objects \\
\midrule
Biology
 & Pump    & syringes (20, 60\,mL) \\
 & Pipette & OniLab pipettes (single-, multi-channel) \\
 & Cap     & dropper bottle; 96-well plate;
             petri dish (50\,mm); glass bottle (250\,mL);
             centrifuge tubes (15, 50\,mL);
             culture tube (14\,mL);
             rectangular bottle (160\,mL);
             reagent bottles (30, 50\,mL) \\
\midrule
Engineering
 & Two-handle  & pliers (1--4) \\
 & Screwdriver & manual screwdriver;
                 electric screwdrivers (large, small) \\
 & Trigger     & impact driver; butane torch; drill \\
\midrule
Home
 & Cap        & marker pen; glue stick; light bulb; child-safe cap; peanut butter jar \\
 & Two-handle & scissors; chopsticks \\
 & Trigger    & squeeze bottle; alcohol bottle \\
 & Pump       & nasal spray \\
 & Reorientation     & kitchen whisk \\
\bottomrule
\end{tabularx}
\vspace{-10pt}
\end{table}

For each object, we executed 10 trials from its prescribed initial configuration using the same parameters across all trials. A trial was considered successful when the intended object operation was completed without manual intervention. The completion criterion depended on the task. Threaded closures were required to reach the specified open or closed state, two-handle tools to complete the commanded opening or closing motion, plungers and triggers to reach the specified displacement, and screwdriving tasks to reach the specified insertion or removal depth. Failure to establish or maintain the required grasp was counted as an unsuccessful trial.

\subsection{Manipulation Across Objects and Tasks}
\label{sec:object_evaluation}

The Cartesian Hand successfully executed all manipulation categories represented in the 35-object evaluation using the primitive library described earlier. Representative examples are shown in Figs.~\ref{fig:objects} and~\ref{fig:sequence}. Twist–translate procedures were used for laboratory bottles, centrifuge and culture tubes, a marker, glue stick, and light bulb. Relative translation between the two grippers was used to operate syringes, pumps, and two-handle tools, while direct gripper actuation was used for spring-loaded triggers. The same primitive library also supported more specialized procedures, including adjustment and actuation of laboratory pipettes and repeated screwdriving motions. Across the 35 objects and 10 trials per object, the hand completed all 350 trials successfully. This high repeatability was supported by the deterministic mapping between commanded joint motion and fingertip displacement, with contact-based grasping that accommodated variations in object dimensions. Given the prescribed initial object pose, the resulting manipulation procedures were highly repeatable across trials.

Figure~\ref{fig:sequence} illustrates how different primitives were composed in a continuous sequence involving three mechanically distinct objects. The hand removed a threaded cap through fingertip twisting, actuated a pipette through relative translation, and pulled a spray-bottle trigger through gripper closure. Although these tasks used different axes and primitive combinations, they were executed using the same underlying primitive library. The results further demonstrate that an all-linear actuation architecture does not restrict the hand to linear object motion. Through grasping, contact interactions, and relative motion between the two grippers, the hand produced both translational and rotational manipulation across different object mechanisms.

\subsection{Reuse Across Objects}
\label{sec:object_adaptation}

We next evaluated how readily the same manipulation procedure could be reused across objects with different geometries. For each manipulation category, we first established the procedure and its primitive sequence using a representative object. New objects within the same category retained this procedure, and only a small set of scalar parameters, such as grasp height, depth offset, stroke distance, force limit, and motion range, were adjusted to account for differences in object geometry. Across all objects evaluated in this work, appropriate parameters for a new object were identified within at most five setup trials.

Several parameters were obtained directly through contact during execution, further reducing the amount of object-specific specification required. For example, when the auxiliary gripper closed around a threaded cap, the final jaw position provided an estimate of the cap diameter, which was then used to determine the fingertip motion and number of repeated strokes required for the desired rotation. Object width was similarly obtained from gripper closure. Parameters that could not be inferred from contact, such as offsets along the fingertip direction or the relative height between object parts, were specified during setup. With the 350 evaluation trials, these results show that the manipulation procedures were reusable across objects within the same category. New objects required parameter adjustment instead of changes to the primitive sequence, controller, or hand mechanism.

\subsection{Humanoid Deployment and Bimanual Manipulation}
\label{sec:humanoid}

We further evaluated the Cartesian Hand on \ifanonymize an in-house developed humanoid \else the Duke Humanoid V2 ~\cite{xia2025duke,xia2026workspace}\fi to examine whether the hand-level manipulation procedures transferred across robot embodiments and could be composed in a bimanual setting. We mounted the same hand on the humanoid and repeated a subset of the manipulation tasks demonstrated on the Franka platform. The hand hardware, primitive definitions, and manipulation procedures remained unchanged. Only the robot-specific approach motion and grasp pose were adapted to the humanoid.

We then equipped both humanoid arms with Cartesian Hands and demonstrated a bimanual laboratory procedure (Fig.~\ref{fig:humanoid-sequence}). One hand grasped a centrifuge tube and opened it independently by holding the tube body with one gripper and twisting the cap with the other, while the second hand operated an adjustable micropipette for liquid transfer. The two arms were used to bring the tube and pipette into the required relative pose, while each hand independently manipulated the object it held. Because each Cartesian Hand provides the opposing grasps and relative motion needed for its own manipulation task, one arm does not need to serve primarily as a fixture for the other. The same single-hand primitives and procedures were retained throughout the bimanual demonstration.

\section{Conclusions, Limitations, and Future Work}
\label{sec:conclusion}

We presented the \texttt{Cartesian Hand}, an all-linear 7-DoF end-effector that combines two independent grasps with relative manipulation within a single hand. Despite its simple kinematics and control, the hand executed a broad range of translational and rotational in-hand dexterous manipulations across 35 laboratory, manufacturing, and household objects, completing all 350 evaluation trials successfully once configured. New objects tested within an established manipulation category required at most five setup trials to determine their parameters, and the same manipulation procedures were applied on a humanoid and composed in a bimanual laboratory task.

The current system is designed for structured manipulation. Its strongest capabilities arise when an object contains a mechanism, such as a thread, pivot, linear guide, plunger, or trigger, that constrains the relative motion between its parts. The experiments assumed prescribed initial object poses, and the hand relied only on joint feedback without directly observing object pose, contact location, or contact force. Future visual and tactile feedback could relax these assumptions while retaining the same mechanical architecture and manipulation capabilities.

More broadly, the \texttt{Cartesian Hand} suggests that greater robotic dexterity does not necessarily require greater mechanical articulation. By combining independent grasping with simple relative motions, the hand turns a small set of directly controlled linear primitives into rich in-hand dexterous manipulation.

\bibliographystyle{IEEEtran}
\bibliography{references}

\end{document}